\documentclass{article}
\usepackage{spconf,amsmath,graphicx}
\usepackage{float}

\title{The Ladder Algorithm: Finding Repetitive Structures in Medical Images by Induction
}
\name{Rhydian Windsor, Amir Jamaludin}
\address{Visual Geometry Group, Department of Engineering, University of Oxford, Oxford, UK}
\begin{document}
\maketitle
\begin{abstract}
In this paper we introduce the Ladder Algorithm; a novel recurrent algorithm to detect repetitive structures in natural images with high accuracy using little training data.

We then demonstrate the algorithm on the task of extracting vertebrae from whole spine magnetic resonance scans with only lumbar MR scans for training data. 
It is shown to achieve high performance with 99.8\% precision and recall, exceeding current state of the art approaches for lumbar vertebrae detection in T1 and T2 weighted scans. It also generalises without retraining to whole spine images with minimal drop in accuracy, achieving 99.4\% detection rate.
\end{abstract}
\section{Introduction}
\label{sec:intro}

Labelling medical images is time-consuming, labour-intensive and boring. This is especially true when detecting or segmenting repetitive structures such as vertebrae, teeth and ribs. Fortunately, we can leverage prior knowledge of spatial relations between instances of repeating objects to perform this task in an automated manner \cite{leung_detecting_1996, hays_discovering_2006}. In this paper we explore a new method of using a priori knowledge of structures such as these to ease the task of detection, resulting in a learning system needing fewer training examples to achieve good performance. 

We demonstrate this algorithm on the task of detecting and labelling vertebral bodies in whole spine magnetic resonance (MR) scans. Extracting vertebrae is an important intermediate task for the automated analysis of spinal diseases and there are several examples of this task being achieved with high accuracy in lumbar-specific MR scans \cite{lootus_vertebrae_2014, Jamaludin15, lu_deepspine:_2018, jamaludin_issls_2017, forsberg_detection_2017}. However, there is very little recent work on extracting vertebrae from whole spine scans. Such scans are important for diagnosing whole spine diseases such as ankylosing spondylitis, bone metastases and multiple myeloma. Most previous approaches for vertebral extraction in the whole spine case do not use deep learning, relying instead on fitting a polynomial to intensity profiles of scans and finding local minima to separate vertebrae \cite{zhigang_peng_automated_2005} or random forest classifiers applied to appearance based features \cite{glocker_vertebrae_2013}. 

Perhaps the reason that deep learning systems have not been popular for MR whole spine vertebral body extraction is a deficiency of labelled data. This may be because many MR scans are lumbar-only as this is the most common site for back pain to occur. Also, whole spine scans take much longer than lumbar scans to annotate, so generating datasets of a similar scale to those available for lumbar scans is difficult.

\begin{figure}[htb]

  \centering
  \centerline{\includegraphics[width=\linewidth]{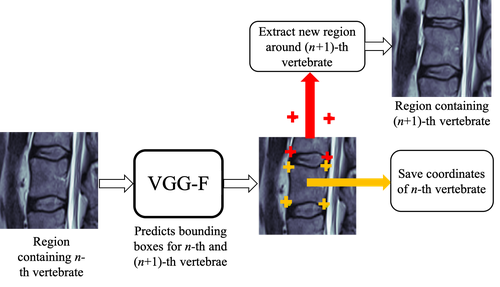}}

\caption{A single iteration of the Ladder Algorithm applied to vertebral body extraction. Given a image patch containing the $n$-th instance of a repetitive object, a network predicts a bounding box around the object (shown by yellow crosses) and proposes a location for the $(n+1)$-th instance (shown by red crosses). This proposal is used to extract the next patch. }
\label{fig:example_iteration}
\end{figure}

As a solution to this problem, we train a model to extract vertebral bodies from lumbar scans in such a way that it will  generalise to whole spine scans. To achieve this, we introduce the Ladder Algorithm; starting at the S1 vertebra at the bottom of the spine each vertebra is bounded iteratively, using the previously detected vertebral body to infer its approximate location. This model is trained using a large cohort of labelled lumbar scans and then applied to whole spine scans simply by increasing the number of iterative predictions made. A single iteration of the algorithm is shown in Figure \ref{fig:example_iteration}.

We evaluate the algorithm by assessing performance on manually labelled 2-D mid-sagittal slices from lumbar and whole spine scans. Following evaluation heuristics previously used in automated vertebral extraction, in particular those of \cite{lootus_vertebrae_2014, forsberg_detection_2017}, the algorithm achieves higher accuracy than state of the art approaches at detecting vertebral bodies in lumbar images. More importantly, the algorithm only has a small drop in accuracy when generalising to whole spine scans and as such is a cheap and data-efficient way to annotate them.
\begin{figure*}[t!]
\centering
\begin{minipage}[b]{1.0\linewidth}
\centering
  \includegraphics[width=0.9\textwidth]{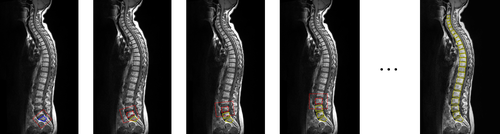}
  \caption{A demonstration of the steps used to detect vertebral bodies for the whole spine. Vertebral detections are made sequentially for a fixed number of iterations up the spine. Here the blue bounding box indicates the coordinates used to initialise the algorithm, the dashed red box indicated proposed regions containing vertebrae and the continuous red box is the patch given as input to the CNN. Yellow boxes indicate saved detections.}
  \label{fig:pipeline}
\end{minipage}
\end{figure*}

\section{THE LADDER ALGORITHM}
\label{sec:overview}

The ladder algorithm is analogous to proof by induction. In inductive proof, the key step is to show that if some property holds for the $n$-th case of a sequence,  it must also hold for the $(n+1)$-th case using knowledge of how the sequence was generated. In the ladder algorithm, given a patch of an image containing the $n$-th instance of some repetitive object we use a convolutional neural network to predict a region containing the location of the $(n+1)$-th instance of the object as well as detecting the $n$-th object from the patch. If we expect $N$ total instances of the object and can accurately extract the $n=1$ case, then detecting all the objects is as simple as applying the algorithm for $N$ iterations. Alternatively, if we do not know how many instances of an object exist in an image then the algorithm can be augmented to include a stopping criterion, identifying when we reach the final object in a sequence and stopping the algorithm.

Detecting objects as a sequence where each prediction is conditioned on previous predictions also has parallels to recurrent neural networks (RNNs) in the case of mapping a single input onto a sequence as an output. However, instead of using a network's hidden state to inform the next prediction we are more explicit, giving a specific patch of the image for the network to use to make its next prediction. The idea of using induction to train RNNs has been explored previously and shown to yield good results at tasks such as text spotting, counting  and segmenting repeating objects \cite{gupta_inductive_2018, romera-paredes_recurrent_2015}.

Having introduced the algorithm, the rest of this paper focuses on its application to the extraction of vertebral bodies from spinal MR scans. However, the algorithm can be applied to much wider range of domains. We suggest some potential medical domains for future work in Section~\ref{sec:discussion}.

\section{EXTRACTING VERTEBRAL BODIES FROM whole spine  MR SCANS}
\label{sec:application_to_spine}

In the experiments in this paper, the ladder algorithm is used to detect bounding boxes around vertebrae in whole spine scans using only vertebrae from lumbar scans as training data. Specifically, we attempt to predict from the S1 vertebra to the C3 vertebra in whole spine scans using annotated sequences of S1 to T12 vertebral bodies in lumbar spine scans as training examples. We do not attempt to detect the C1 and C2 vertebral bodies as they appear very different to the rest of the vertebrae in the spine.

An example of the algorithm being used to localise vertebral bodies in a whole spine scan is shown in Figure \ref{fig:pipeline}. Firstly, a detector trained on extracting vertebrae from lumbar MR scans such as \cite{lootus_vertebrae_2014} is used to extract the location of the S1 vertebra from a whole spine scan. This bounding box is used to extract a patch surrounding the S1 vertebral body which is then input to the network shown in Figure \ref{fig:example_iteration}. This predicts co-ordinates in the patch for vertices of bounding boxes for the S1 vertebra and the vertebra above it, L5 (details on how the network is trained are given in Section \ref{sec:architecture_and_training}). The resulting image coordinates of S1 are saved as final predictions whereas the coordinates of the L5 are used to extract a new patch of the image, centred on these coordinates. This new patch is then used as a new input to the network, which extracts the final coordinates of L5 and proposed coordinates for the body of the next vertebra, L4. This process repeats iteratively up the spine for a fixed number iterations. In the experiments in this paper, 23 iterations are used for full spine scans, resulting in detection of all vertebrae from S1 to C3 in almost all cases. However, an estimated 7.7\% of the population have variations in the number of vertebrae in the spine \cite{tins_incidence_2016}, meaning the exact range of vertebrae found cannot be determined trivially.  The impact of this is discussed in Section \ref{sec:discussion}.

\subsection{Datasets Used}
\label{sec:datasets}
As stated, two datasets are used for the experiments in this paper. \textbf{Genodisc}  is a large dataset consisting of 5404 T1 and T2 weighted MR lumbar scans from 2295 unique patients collected by the Genodisc consortium. %
Each scan has ground truth bounding boxes for S1, L1-5 and T12, extracted by the method outlined by Lootus \textit{et al}.\ \cite{lootus_vertebrae_2014} and manually checked for accuracy. In this experiment, Genodisc is randomly split into training, validation and test sets with splits of 80\%, 10\% and 10\% respectively. Splits were done to ensure that all scans of each patient remained in the same set. 

Oxford Whole Spine (\textbf{OWS}) is a dataset of 64 T1 and T2 weighted 3-D clinical whole spine MR scans from unique patients. The cohort consists of a range of degenerative changes including scoliosis, stenosis and collapsed vertebrae. These scans are manually annotated with bounding boxes around each of the 23 vertebrae from S1 to C3. OWS is used here solely as a testing dataset. 

\begin{figure}[htb]

\begin{minipage}[b]{1.0\linewidth}
  \centering
  \centerline{\includegraphics[width=8.cm]{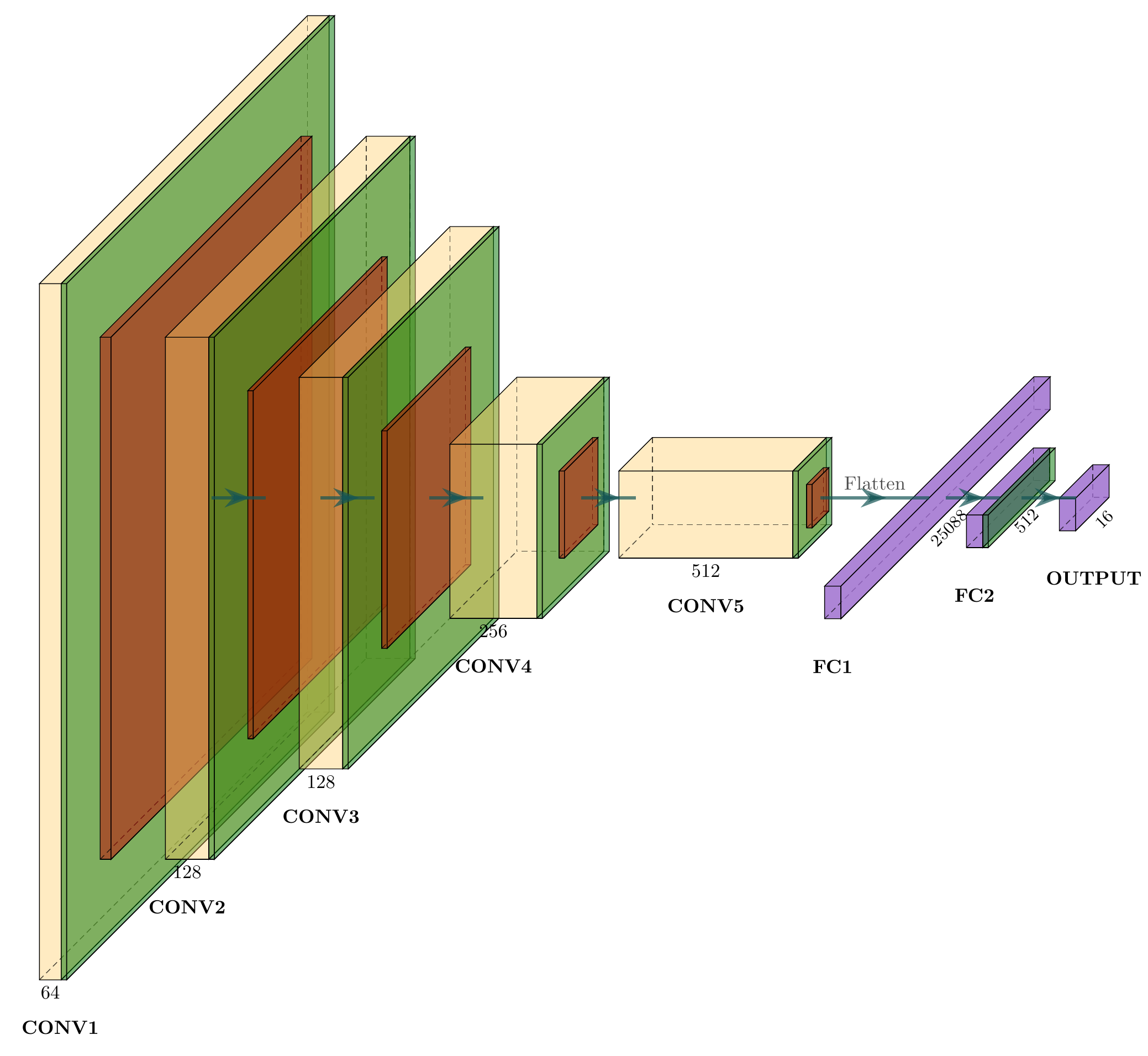}}
  \caption{The architecture used recurrently to extract vertebral corners. Here yellow blocks are 2-D convolutional layers, green blocks represent batch normalisation operations and red blocks represent max pooling operations. Purple blocks show fully connected layers. Each convolutional layer has a stride of (1,1), a padding size of (1,1) with a kernel size of $(3,3)$ and is followed by ReLU activation units (not shown). Inputs are $224\times224$ single channel patches, as shown in Figure \ref{fig:example_iteration}.  }
  \label{fig:arch}
\end{minipage}

\end{figure}

\subsection{Network Architecture and Training}
\label{sec:architecture_and_training}

Instead of directly detecting bounding boxes we predict the corners of vertebral bodies, specifically the 4 corners of the central vertebral body in the patch and the 4 corners of the vertebral body above. To predict the corners, a variant on the VGG-F architecture \cite{chatfield_return_2014} is used. However, unlike the initial formulation, batch normalisation is used between layers and each convolutional layer uses a kernel size of (3,3). A diagram of the architecture used is shown in Figure \ref{fig:arch}. The 16 outputs are the x and y coordinates of the 8 vertebral corners; four corners for the central vertebral body and four corners for the vertebral body above it. We choose to regress exact co-ordinates of vertebral body corners rather than the normal approach in object detection using deformable part models \cite{Felzenszwalb_dpm_2008} of regressing heatmaps to the corner landmarks \cite{jain_learning_2013, tompson_joint_2014, pfister_flowing_2015, payer_regressing_2016}. This is done so that the network can predict corners not visible in the patch, which is not possible with the heatmap approach.

The network is trained using mid-sagittal slices from the lumbar scans in the Genodisc training set. Each training example consists of a patch centred on a bounding box for a vertebra from S1 to L1. The coordinates for the bounding boxes of the central vertebra and the one above it are provided as labels. This patch is extracted by drawing a tightly fitting rectangle around the bounding box and then extending it by 75\% on each side. Cubic interpolation is then used to resize the patch to be $224\times224$ pixels. This ensures constant input tensor size. Ground truth labels are extracted by applying the same transformations to the ground truth bounding box co-ordinates in the original image. The network is tasked with predicting these transformed co-ordinates using the patch as input. It is trained by minimising the L2 loss between the output vector and a 16-dimensional vector consisting of the x and y coordinates of the ground truth corners for both vertebral bodies.

The network is trained using an Adam optimiser \cite{kingma_adam_2015} with $\beta_1=0.9$, $\beta_2=0.999$ and a learning rate of $10^{-4}$ until convergence. Training augmentation is applied by translating the patch used by $\pm$10 pixels in the original image, expanding the rectangle on each side by a random amount sampled uniformly from 60\% to 90\% and by flipping the patches in the coronal (vertical) plane. Vertebrae decrease in size as we move up the spine so a Gaussian blur with covariance randomly selected from 3 to 30 pixels is applied to the extracted $224\times224$ patch. This mimics blur from resizing the relatively smaller cervical vertebral bodies at the top of the spine.

\section{PERFORMANCE EVALUATION \& RESULTS}
\label{sec:evaluation}

\subsection{Evaluation}
The algorithm is assessed on both lumbar and whole spine scans from the testing set of Genodisc and OWS respectively. In both cases, the algorithm was initialised by extracting the S1 vertebra in an entirely automated manner using a method derived from \cite{lootus_vertebrae_2014} and applied for 6 iterations in lumbar scans and 23 for whole spine scans. Following \cite{forsberg_detection_2017}, a true positive detection is defined as when the centroid of a predicted bounding box is inside a ground truth vertebral body, a false positive detection as when the centroid is not inside a vertebral body and a false negative detection as when a ground truth bounding box contains no predicted centroids. 

To demonstrate the ability of the algorithm to regress bounding box corner coordinates accurately we report Dice coefficients for the overlap between predicted bounding boxes and ground truth boxes in correct detection cases.

\begin{figure}[h!]
  \centering
  \centerline{\includegraphics[width=7.5cm]{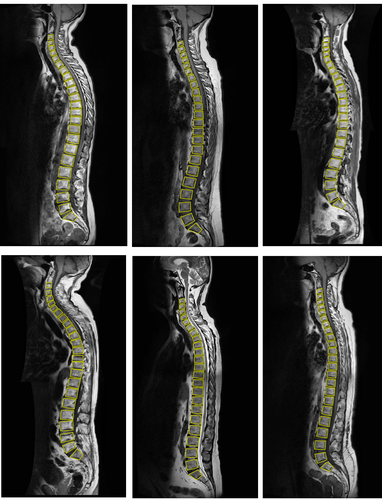}}
  \caption{Example vertebral extractions from mid-sagittal slices of whole spine scans in 6 different patients.}
  \label{fig:example_results}
\end{figure}

\subsection{Results}

The recall and precision of the classifier are shown in Table \ref{table:results} for both the whole spine (WS) and lumbar (L) spine cases. Figure \ref{fig:example_results} shows the examples of the algorithm applied to full spine scans.

\begin{table}[!h]
\setlength{\tabcolsep}{5pt}
\centering
{\footnotesize
\begin{tabular}{l l l c c} 

    \multicolumn{1}{l}{Scan Type} & \multicolumn{1}{c}{Recall(\%)} & \multicolumn{1}{c}{Precision(\%)} & Dice & LE (mm) \\
 \hline
WS (All) & 99.4 [1399/1408] & 99.4 [1399/1408] & 91.8 & 0.98 $\pm$ 0.68\\ 
WS (T1) & 99.2 [982/990] & 99.2 [982/990] & 91.6& 1.00 $\pm$ 0.80\\ 
WS (T2) & 99.8 [417/418] & 99.8 [417/418] & 91.9& 0.93 $\pm$ 0.61\\ 
L (All) & 99.8 [2872/2878] & 99.8 [2871/2878] & 93.5 & 0.62 $\pm$ 0.85\\ 
L (T1) & 99.6  [1344/1349]& 99.6 [1344/1349] & 93.2 & 0.65 $\pm$ 0.99\\ 
L (T2) & 99.9 [1528/1529] & 99.9 [1528/1529] & 93.8 & 0.58 $\pm$ 0.71 \\ 

\end{tabular}
}
\caption{Performance of the classifier on both whole spine (WS) and lumbar (L) scans from OWS and Genodisc's test sets respectively. Results are reported on a vertebral level. The final column shows the centroid localisation error (LE).}
\label{table:results}
\end{table}

\section{DISCUSSION \& CONCLUSION}
\label{sec:discussion}

In this paper we have presented the Ladder Algorithm, a method of extracting repetitive structures from images with little training data. We have demonstrated this algorithm by using it to localise vertebral bodies in whole spine images with only annotated lumbar scans as training data.

This algorithm compares favourably to state of the art approaches at detecting vertebral bodies in lumbar scans, achieving detection rates 99.6-99.9\%. For comparison, the current highest accuracy reported at this task is that of Forsberg \textit{et al}.\ \cite{forsberg_detection_2017}, which achieves 99.3-99.6\% accuracy. We also report a lower localistation error of vertebrae centroids. However, it should be noted that these evaluations are done on different datasets and thus do not reflect a true comparison.

Encouragingly, the performance of this algorithm only decreases slightly when applied to whole spine scans, achieving 99.2-99.8\% sensitivity and specificity. Failure cases in this example are generally due to partially visible vertebrae caused by the scanner being misaligned with the spine's orientation.

There are two major drawbacks to this approach; the algorithm runs for a fixed number of iterations meaning it is not adaptive to variations in the number of vertebrae in the spine and it relies on initialising the location of the S1 vertebra using another algorithm. Rectifying these issues is left as future work although we suggest that one solution could be to use this system to rapidly label a large amount of whole spine data and use this for training a new system robust to variable numbers of vertebrae.

Future work on this will include demonstrating the algorithm generalises to different modalities, for example CT and PET scans as well as  to different tasks including teeth and rib detection.

\section{ACKNOWLEDGEMENTS}
\label{sec:acknowledgements}
The authors would like to thank Timor Kadir and Andrew Zisserman for useful discussion and feedback during the researching of this paper. Rhydian Windsor is supported by Cancer Research UK as part of the EPSRC CDT in Autonomous Intelligent Machines and Systems (EP/L015897/1).
Amir Jamaludin is supported by EPSRC Programme Grant Seebibyte  (EP/M013774/1). The data was obtained during the ECFP7 project GENODISC (HEALTH-F2-2008-201626).

\bibliographystyle{IEEEbib}
\bibliography{refs}

\end{document}